\documentclass{article}
\usepackage{iclr2027_conference,times}
\usepackage[utf8]{inputenc}
\usepackage[T1]{fontenc}
\usepackage{amsmath,amssymb,booktabs,array,tabularx,microtype}
\usepackage{graphicx}
\usepackage[table]{xcolor}
\usepackage[hidelinks]{hyperref}
\usepackage{enumitem}
\usepackage{cleveref}
\usepackage{float}
\usepackage{tikz}
\usepackage{wrapfig}

\definecolor{accent}{HTML}{254F72}
\definecolor{tint}{HTML}{EDF3F7}
\definecolor{lightgrey}{HTML}{F2F3F4}
\definecolor{revisionblue}{HTML}{0057B8}
\definecolor{revisionyellow}{HTML}{B8860B}
\DeclareRobustCommand{\rev}[1]{\textcolor{revisionblue}{#1}}
\DeclareRobustCommand{\revy}[1]{\textcolor{revisionyellow}{#1}}
\definecolor{suffbg}{HTML}{E8F5E9}
\definecolor{sufffg}{HTML}{2E7D32}
\definecolor{absentbg}{HTML}{FFEBEE}
\definecolor{absentfg}{HTML}{C62828}
\definecolor{partialbg}{HTML}{FFF3E0}
\definecolor{partialfg}{HTML}{E65100}
\definecolor{conflbg}{HTML}{F3E5F5}
\definecolor{conflfg}{HTML}{6A1B9A}

\newcommand{\evpill}[3]{%
  \tikz[baseline=(t.base)]{%
    \node[fill=#1,text=#2,rounded corners=2pt,
      inner sep=1.5pt,inner xsep=3pt] (t)
      {\textsc{\footnotesize #3}};%
  }%
}
\newcommand{\suff}{\evpill{suffbg}{sufffg}{sufficient}}
\newcommand{\partl}{\evpill{partialbg}{partialfg}{partial}}
\newcommand{\absent}{\evpill{absentbg}{absentfg}{absent}}
\newcommand{\confl}{\evpill{conflbg}{conflfg}{conflicting}}

\newcommand{\ind}{\mathbf{1}}

\title{EviDial: When Correct Depression Predictions Lack Evidence}

\author{
  \textbf{
    Junjie Xu\textsuperscript{1},
    Jin Zhu\textsuperscript{1},
    Yuzhe Yang\textsuperscript{1},
    Luhan Wang\textsuperscript{1},
    Xingjiao Wu\textsuperscript{1},}\\
  \textbf{
    Daoguo Dong\textsuperscript{2},
    Jing Yang\textsuperscript{1},
    Wenxin Hu\textsuperscript{1},
    Liang He\textsuperscript{1}}\\
  \textsuperscript{1}East China Normal University\\
  \textsuperscript{2}Fudan University
}

\iclrfinalcopy

\begin{document}
\maketitle
\lhead{}

\renewcommand{\topfraction}{0.9}
\renewcommand{\textfraction}{0.07}
\raggedbottom

\begin{abstract}

Depression dialogue benchmarks typically evaluate models by comparing predicted symptom labels against clinical scale scores. Yet a prediction can match the benchmark label even when the dialogue does not contain sufficient evidence to support that assessment. We introduce \textsc{EviDial}, a benchmark for measuring this distinction between \emph{label agreement} and \emph{evidence-grounded assessment}. \textsc{EviDial} covers 382 depression dialogues from three sources and 2,744 participant--symptom cells, each characterized under an evidence taxonomy of \emph{absent}, \emph{partial}, \emph{conflicting}, or \emph{sufficient} evidence.
Because evidence sufficiency itself has no directly observable gold standard, we construct calibrated evidence references from multiple judgments using generalizability theory, many-facet Rasch measurement, and random-effects latent class analysis. Auditing existing benchmarks with these references reveals a substantial evidence gap: 82.8\% of participant--symptom cells do not contain sufficient textual evidence for the corresponding symptom assessment, with evidence availability varying sharply across data sources. More importantly, across evaluated model families and prompting strategies, 69.2--96.9\% of predictions that are \emph{correct according to the benchmark label} remain unsupported by sufficient evidence in the dialogue. Additional measurement analyses show that high agreement among LLM evaluators does not necessarily imply independent evidence judgments, motivating explicit calibration of automated reference panels.
Conventional label-based evaluation can substantially overstate evidence-grounded performance by conflating agreement with clinical labels and support from the observed dialogue. \textsc{EviDial} provides a measurement layer for separating these two constructs and enables more faithful auditing of LLM behavioral assessment in depression dialogues.

\end{abstract}

\section{Introduction}\label{sec:intro}

\begin{figure}[t]
\centering
\includegraphics[width=0.9\textwidth]{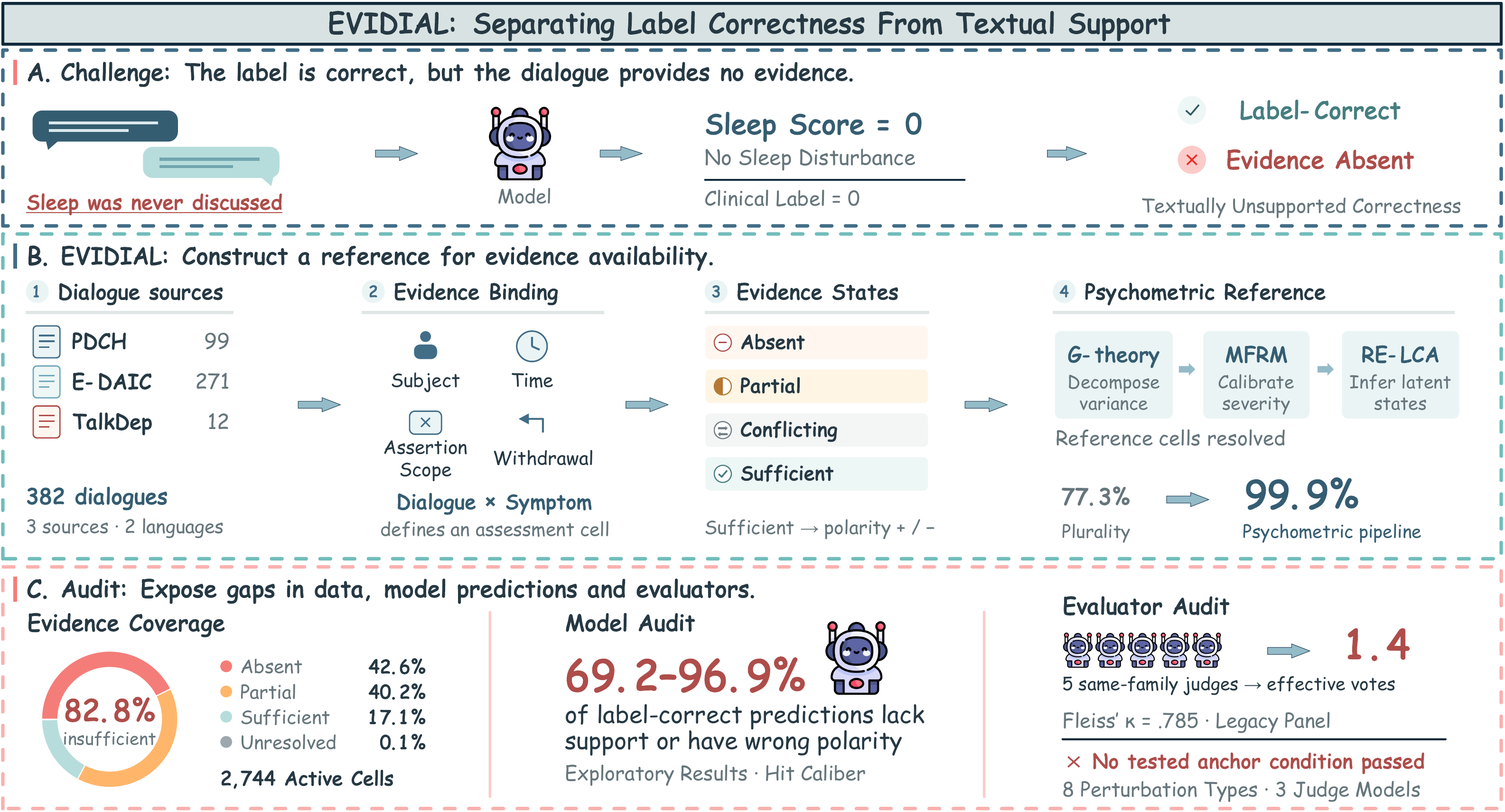}
\vspace{-12pt}
\caption{Existing depression dialogue benchmarks have one \textbf{challenge}: \revy{Textually Unsupported Correctness} \textit{(A.)}. We propose EVIDIAL, a reference benchmark for evidence availability \textit{(B.)}, and audit this phenomenon by exposing gaps in data, model predictions, and evaluators \textit{(C.)}.}
\vspace{-12pt}
\label{fig:overview}
\end{figure}

Depression dialogue benchmarks record real or simulated conversations between clinicians and individuals with potential depression, and often evaluate models by comparing predicted symptom labels with clinical or self-report scores.
As shown in Figure \ref{fig:overview} A., consider an interview paired with a recorded sleep-symptom label.
A model is asked to evaluate the respondent on the sleep disturbance dimension.
It predicts \emph{no sleep disturbance}-and matches the native symptom label exactly.
However, label accuracy is a success: one more correct cell in the accuracy table.
But the interview never discussed sleep.
The model produced the right answer without supporting evidence in the released dialogue--one label--support divergence we term \textbf{\revy{Textually Unsupported Correctness}}. The term locates a mismatch between a credited prediction and the observable input; it is not a mechanism claim that the model ignored evidence, used priors, or reasoned in any particular way.

Benchmarks such as PDCH \citep{cao2025pdch}, E-DAIC \citep{gratch2014daic}, EATD \citep{shen2022eatd}, and their derivatives report label accuracy, macro F1, or mean absolute error.
These metrics do not directly distinguish two fundamentally different outcomes: \rev{a prediction grounded in textual evidence and one that matches the label without support for the predicted symptom state}.
An explicit denial and silence about a symptom can receive the same negative label, although only the denial provides negative evidence within its stated scope.
Label agreement alone therefore cannot establish that an assessment is justified by the available dialogue.
A native clinical score may incorporate information outside the released transcript, and the issue here is \rev{what the text supports, rather than whether that score is wrong}.

We introduce \textsc{EviDial} (Evidence-Dialogue), a benchmark designed to make this distinction explicit.
Across 2,744 assessment cells from 343 clinical participants, with eight source-specific dimensions per participant, 82.8\% of cells are assigned absent or partial evidence by the candidate reference (95\% CI: 82.1--83.5\%).
This pooled share is composition-dominated: sufficient evidence is assigned to 74.1\% of PDCH cells but only 1.9\% of E-DAIC cells, so \textbf{evidence availability behaves as a property of the benchmark--instrument--transcript interface rather than of the clinical label itself}.
The gap is also not confined to negative labels, where negative evidence is structurally harder to establish (\S\ref{sec:taxonomy}): 76.0\% of positive-native-label cells lack sufficient evidence (95\% CI: 73.9--78.1\%), against 91.7\% of negative-label cells.
Among predictions that do match native labels, 69.2--96.9\% lack reference support for the predicted direction across three model families and four prompting methods on 87 development cells.
As a fraction of all planned predictions, the corresponding incidence is 20.7--54.0\%---a quantity we term the \revy{unsupported-correct rate} (UCR).
These findings motivate an audit of both model predictions and the evidence judgments used to evaluate them.
\revy{Textually Unsupported Correctness} matters when benchmark accuracy is interpreted as evidence of clinical assessment capability.
If a model achieves high accuracy on depression severity without an audit of textual support, we cannot tell which credited predictions are warranted by its input.
A model that has learned ``most people in this corpus score 0 on sleep disturbance'' could receive credit without establishing that a particular respondent denied the symptom~\citep{geirhos2020shortcut}.
If that regularity changes in another population, the original accuracy score alone gives no assurance that those predictions will remain reliable.
Evaluation therefore needs to distinguish supported predictions, unsupported label matches, and cases in which the available dialogue warrants abstention or further inquiry.
Making these differences visible clarifies what benchmark performance can substantiate before it is used to argue for clinical readiness.

\textsc{EviDial} adds this evidence-support layer to label-based evaluation.
The core collection comprises 382 dialogues from three sources: PDCH (99 dialogues, Chinese), E-DAIC (271 dialogues, English), and TalkDep (12 dialogues, English).
Repeated PDCH visits and simulated TalkDep dialogues are outside the main clinical reference pool.
Each dialogue--dimension pair constitutes an \emph{evidence cell}, annotated under a four-state taxonomy: \absent{} (no eligible evidence), \partl{} (some indicators present but insufficient for the scoped binary target), \confl{} (contradictory signals), or \suff{} (enough evidence to support one target polarity).
Annotations are grounded through formal binding operations that enforce constraints on subject identity, temporal reference, negation scope, and symptom withdrawal over judge-extracted evidence frames, which makes the distinction between missing evidence and evidence of symptom absence explicit in the evaluation target.

Since no single annotator or rubric can serve as an uncontested gold standard~\citep{artstein2008agreement}, we audit how the evidence reference is constructed.
Agreement among judges does not by itself establish that their evidence assignments are valid.
Random-effects latent class analysis (RE-LCA)~\citep{qu1996lca} combines the new panel's categorical votes into a model-consensus candidate reference.
Separately, generalizability theory (G-theory)~\citep{brennan2001gtheory} quantifies variance components across cells, participants, and protocols in a legacy panel; many-facet Rasch measurement (MFRM)~\citep{linacre1989mfrm} calibrates legacy protocol severity.
Aggregation comparisons expose sensitivity to reference choices, while perturbation anchors test judgments against controlled changes to evidence.
Our findings remain conditional on the candidate reference; a blinded human validation study (\S\ref{sec:Human-validation}) supports the main evidence gap, though uncertainty calibration is incomplete.
Our work makes \underline{three contributions}:
\noindent\textbf{1. Evidence-support taxonomy and benchmark}.
We introduce \textsc{EviDial}, which augments conventional symptom-label evaluation with a four-state taxonomy of textual evidence support, enabling direct assessment of whether a prediction is justified by the dialogue.
\noindent\textbf{2. Psychometric reference construction and audit}.
We develop a psychometric framework for constructing and auditing evidence references when no single judge can be treated as ground truth, explicitly accounting for judge dependence and reference uncertainty.
\noindent\textbf{3. Empirical evidence-gap audit.}
Using \textsc{EviDial}, we show that label correctness and evidence support can diverge substantially across models and prompting strategies, revealing failure modes that conventional accuracy alone cannot capture.

\vspace{-10pt}
\section{Related Work}\label{sec:related}
\vspace{-10pt}
\noindent\textbf{Depression and psychiatric dialogue benchmarks.}
Clinical dialogue resources such as DAIC~\citep{gratch2014daic}, PDCH~\citep{cao2025pdch}, and EATD~\citep{shen2022eatd} support depression assessment from interviews or elicited speech, while newer benchmarks examine robustness across splits, models, and symptom distributions~\citep{wang2026multiprobe}. Broader psychiatric benchmarks evaluate diagnostic capabilities of LLMs and multi-agent systems~\citep{yang2026mentalbench,hou2026lingxi}. These benchmarks primarily evaluate agreement with clinical labels or diagnostic targets. \textsc{EviDial} instead asks a complementary question: whether the processed dialogue itself contains sufficient evidence to justify the assessed symptom label.

\noindent\textbf{LLM evaluators and reference construction.}
LLMs are increasingly used as evaluators~\citep{zheng2023mtbench,zhao2024judgebench,kim2024prometheus,kc2026babeljudge}, including multi-judge panels that can improve reliability over individual judges~\citep{verga2024juries}. However, correlated errors and shared model biases can make apparent agreement overstate the amount of independent information in a panel~\citep{kohli2026correlated}. Recent work therefore applies psychometric tools such as item-response theory, generalizability theory~\citep{brennan2001gtheory}, and Rasch measurement~\citep{linacre1989mfrm} to characterize LLM-rater reliability and severity~\citep{choi2026irt,wang2025gtheory,song2025rasch,sunkavalli2026rater}. Separately, latent-class and observer-error models estimate reference states when ground truth is unavailable~\citep{dawid1979em,qu1996lca,albert2004lca}. We bring these perspectives together to construct and audit evidence references without treating either a single LLM judge or simple panel agreement as ground truth.

\noindent\textbf{Evidence support, faithfulness, and selective prediction.}
Our distinction between label correctness and textual support is related to faithfulness and attribution in generation~\citep{maynez2020faithfulness,rashkin2023attribution,gao2023rarr,honovich2022true,min2023factscore}, where an output may appear correct while lacking support from its source. Clinical NLP has likewise modeled contextual properties such as negation, temporality, and experiencer~\citep{chapman2001negex,harkema2009context}, while work on context sufficiency and selective prediction studies whether available information is adequate for answering and when a model should abstain~\citep{joren2025sufficient,geifman2017selective,kamath2020selective}. \textsc{EviDial} adapts these ideas to symptom-level behavioral assessment by explicitly measuring whether benchmark-correct predictions are supported by the evidence available in the dialogue.

\vspace{-10pt}
\section{The EviDial Benchmark}\label{sec:benchmark}
\vspace{-10pt}
To audit the gap between label correctness and evidence-grounded assessment, \textsc{EviDial} defines an assessment unit, an evidence taxonomy with explicit binding rules, a multi-judge reference construction procedure, and evaluation metrics that separate label agreement from textual support.

\subsection{Assessment Cells and Data Sources}\label{sec:data}

The unit of analysis is an \emph{assessment cell} $u=(x,d)$, pairing a dialogue transcript $x$ with a symptom dimension $d$ and asking whether the transcript contains evidence bearing on that symptom for the participant.
The benchmark contains 382 depression-dialogue transcripts from three sources. \textbf{PDCH} contains 99 Chinese psychiatric interviews from 72 participants, including 27 repeated visits, scored with HAMD-17~\citep{hamilton1960hamd}. \textbf{E-DAIC} contains 271 English semi-structured distress interviews scored with PHQ-8~\citep{gratch2014daic,kroenke2009phq8}. \textbf{TalkDep} contains 12 scripted dialogues generated from BDI-II persona profiles; because these profiles are construction targets rather than independent clinical ground truth, TalkDep is excluded from the main reference pool.
We align the sources to nine symptom dimensions: depressed mood, anhedonia, sleep disturbance, fatigue/energy, appetite/weight change, psychomotor change, guilt/worthlessness, concentration difficulty, and suicidal ideation. PDCH excludes concentration and E-DAIC excludes suicidality, yielding $72\times8+271\times8=2{,}744$ main-reference cells. Participants are split without overlap into 182 development and 161 held-out test participants; the model audit in \S\ref{sec:ucr} uses 87 development cells from 76 participants.
E-DAIC transcripts were reformatted with \texttt{claude-3-5-sonnet-latest} (271/275 succeeded). Because preservation of clinical meaning has not been independently verified, our audit concerns the processed model-visible input rather than guaranteed coverage of the original interviews. Appendix~\ref{app:sources} provides source statistics.

\begin{table}[t]
\centering
\vspace{-10pt}
\caption{Binding operations applied to a sleep-disturbance cell. Each row shows a candidate evidence frame from the dialogue, the binding outcome, and its effect on the eligible evidence set.}
\label{tab:binding-coverage}

\scriptsize
\setlength{\tabcolsep}{3pt}
\renewcommand{\arraystretch}{0.92}

\begin{tabularx}{\columnwidth}{@{}X
>{\raggedright\arraybackslash}p{0.16\columnwidth}
>{\raggedright\arraybackslash}p{0.25\columnwidth}
>{\raggedright\arraybackslash}p{0.17\columnwidth}@{}}
\toprule
Dialogue excerpt & Binding test & Result & Status \\
\midrule
``My mother can't sleep at all''
& Subject
& Other person
& Excluded \\

``I had terrible insomnia in college''
& Temporal
& Outside window
& Excluded \\

``I haven't been sleeping well this week''
& All pass
& Affirms sleep difficulty
& Eligible ($+$) \\

``Well, actually I meant last month''
& Withdrawal
& Retracts prior frame
& Prior removed \\
\midrule
\multicolumn{4}{@{}l@{}}
{$\mathcal{E}(u)=\emptyset$ after withdrawal $\Rightarrow s(u)=$ \absent{}} \\
\bottomrule
\end{tabularx}
\end{table}

\subsection{Evidence Taxonomy and Binding Operations}\label{sec:taxonomy}

A prediction is evidence-grounded only if the dialogue contains eligible evidence for the target symptom. Candidate evidence frames are filtered by four binding operations: \textbf{subject binding} requires that the statement concern the participant; \textbf{temporal binding} restricts evidence to the instrument's target window; \textbf{assertion scope} determines whether the statement covers the relevant symptom components; and \textbf{withdrawal} removes subsequently retracted evidence. Presence requires at least one qualifying component, whereas establishing absence requires coverage of all relevant components, making sufficient-negative evidence structurally harder to establish. We audit this asymmetry separately in \S\ref{sec:measurement} and \S\ref{sec:landscape}.

After binding, the surviving evidence frames $\mathcal{E}(u)$ determine compatible binary target values $K(u)\subseteq\{0,1\}$:
\begin{equation}\label{eq:evidence-state}
s(u)=
\left\{
\begin{array}{ll@{\qquad}ll}
\absent{}, & \mathcal{E}(u)=\emptyset,
&
\confl{}, & \mathcal{E}(u)\ne\emptyset,\ K(u)=\emptyset,
\\[2pt]
\suff{}, & \mathcal{E}(u)\ne\emptyset,\ |K(u)|=1,
&
\partl{}, & \mathcal{E}(u)\ne\emptyset,\ K(u)=\{0,1\}.
\end{array}
\right.
\end{equation}
For sufficient evidence, $K=\{1\}$ and $K=\{0\}$ denote positive and negative polarity, yielding five joint reference categories. An \absent{} state means that no eligible evidence survives binding, not necessarily that the symptom was never mentioned. Table~\ref{tab:binding-coverage} gives a concrete example.

\subsection{Reference Construction}\label{sec:reference}
Because evidence state has no directly observable gold-standard annotator, we treat reference construction as a measurement problem with competing estimators. On the same vote pool, we compare strict plurality, a one-hot mean baseline, and random-effects latent-class fusion; \S\ref{sec:measurement} reports their coverage and validation diagnostics.
The working model-consensus reference uses judgments from \texttt{gemini-2.5-flash}, \texttt{qwen3.8-max}, and \texttt{claude-sonnet-4-6}, stratified by language. Two DeepSeek models were excluded after failing format-validity gates. The retained pool contains 5,240 valid judgments over 2,744 cells, with uneven per-cell coverage (Table~\ref{tab:reference-panel-coverage}).
We fuse these judgments using random-effects latent class analysis (RE-LCA)~\citep{qu1996lca,albert2004lca}, with five latent evidence states, judge-specific emission probabilities, and a cell-level random effect capturing dependence. RE-LCA serves as the working reference because it resolves split votes, not because it is assumed uniquely correct: available anchor comparisons favor plurality on correctness, so aggregation choice remains an explicit sensitivity analysis. Anchor-derived emission priors and the uncalibrated posterior-abstention threshold are examined in \S\ref{sec:measurement}; Appendix~\ref{app:reference-fusion} gives the full specification.

\subsection{Evaluation Protocol}\label{sec:metrics}

The native binary target is
$Y=\ind[\text{item score}>0]$
under the source instrument. Because native scores may incorporate information unavailable in the transcript, disagreement between label and evidence does not imply that the clinical assessment itself is incorrect.
Our primary metric is the \textbf{Unsupported-Correct Rate (UCR)},
\begin{equation}\label{eq:ucr}
\mathrm{UCR}=\frac{U+D}{N-R},
\end{equation}
where $U$ counts label-correct predictions whose reference evidence is absent, partial, or conflicting; $D$ counts label-correct predictions for which sufficient evidence supports the opposite direction; $N$ is the number of planned predictions; and $R$ counts nonfailed predictions with unresolved references. UCR is therefore an incidence per planned prediction, not the unsupported proportion among correct predictions.
\begin{wraptable}[8]{r}{0.37\columnwidth}
\vspace{-17pt}
\centering
\caption{Judge coverage by language stratum. Fractions denote valid judgments over all cells.}
\vspace{5pt}
\label{tab:reference-panel-coverage}
\scriptsize
\setlength{\tabcolsep}{2.5pt}
\renewcommand{\arraystretch}{0.95}
\begin{tabular}{@{}lrr@{}}
\toprule
Judge & Eng. ($n=2{,}168$) & Chi. ($n=576$) \\
\midrule
Qwen & 2,167/2,168 & 572/576 \\
Gemini & 1,666/2,168 & --- \\
Claude & 568/2,168 & 267/576 \\
\bottomrule
\end{tabular}
\end{wraptable}
We additionally report \textbf{hit caliber} $(U+D)/(C-R_c)$, the unsupported fraction among resolved correct predictions, together with sensitivity bounds, abstention rate, failure rate, and answered coverage. These quantities are reported jointly because abstention or technical failure can reduce N-based UCR without improving support. Because sufficient-negative evidence is structurally harder to establish, \S\ref{sec:landscape} also reports label-conditional corpus-level gaps.

\noindent\textbf{Models and methods.}
We evaluate Gemini-2.5-flash-lite, GPT-5-mini, and Qwen-plus on 87 development cells from 76 participants under four prompting strategies, with one prediction per model--method--cell. \textbf{B0} directly predicts symptom presence with optional abstention. \textbf{B1} additionally requires a decisive patient quote and checks subject, time window, and scope. \textbf{B2} pairs forced binary prediction with a sufficient-context gate, abstaining when context is insufficient. \textbf{M} jointly elicits evidence frames, evidence state, and polarity, abstaining for non-sufficient states. Results are exploratory; prospective held-out evaluation with frontier models and repeated runs is ongoing.

\vspace{-10pt}
\section{Results}\label{sec:results}
\vspace{-10pt}
\begin{wrapfigure}[10]{r}{0.35\textwidth}
\centering
\vspace{-15pt}
\includegraphics[width=0.35\textwidth]{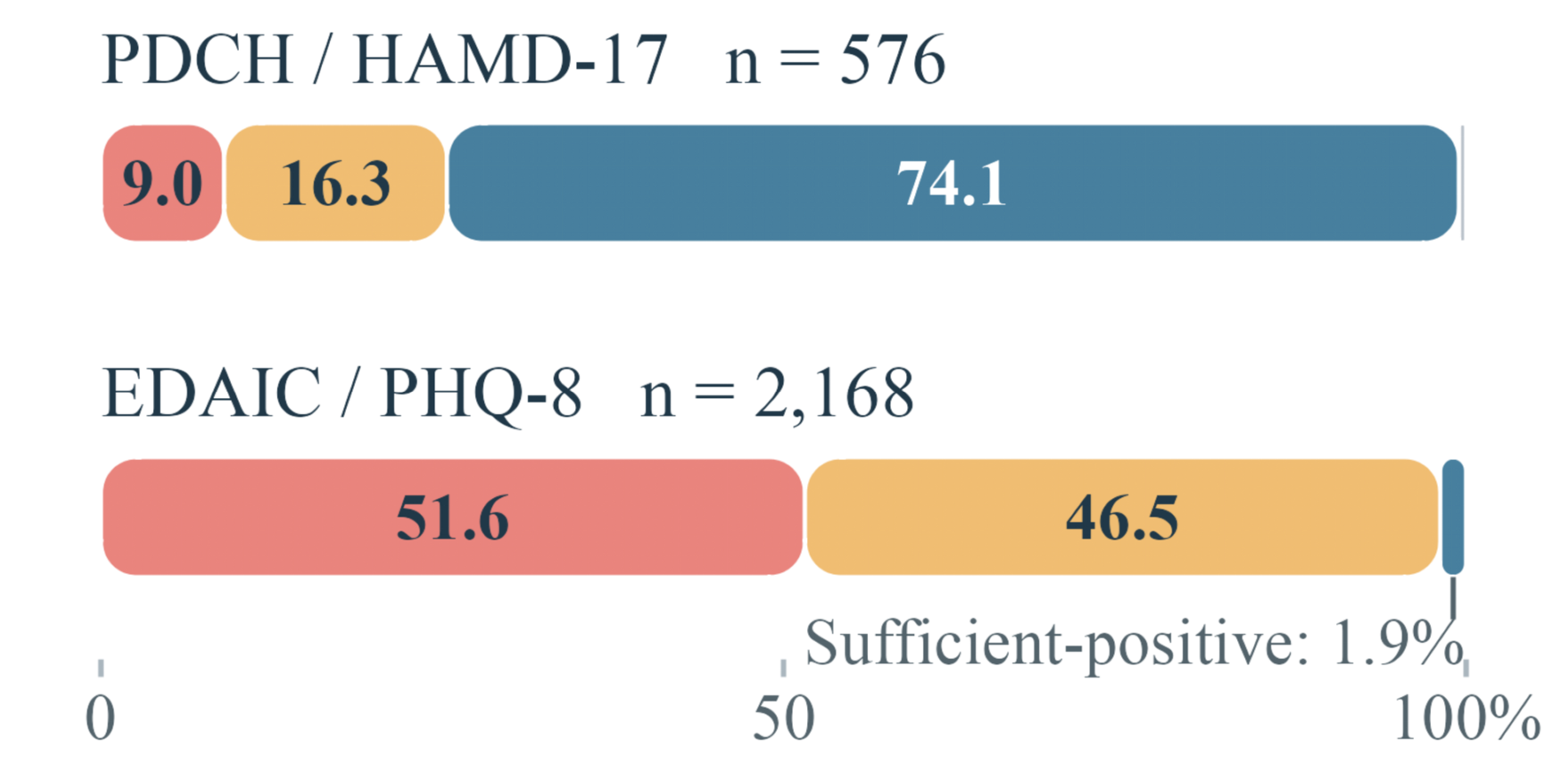}
\vspace{-17pt}
\caption{Within-source evidence composition. All source cells in each denominator.}
\label{fig:within-source}

\end{wrapfigure}
We present findings as a progressive audit in three layers. We begin by establishing the scale of the evidence gap: how much of the benchmark actually contains sufficient textual evidence for the symptom assessments it evaluates (\S\ref{sec:landscape})? The answer---82.8\% of cells lack sufficient evidence---reframes every subsequent accuracy number. We then turn to model evaluation: given this evidence landscape, what does a ``correct'' prediction actually reflect (\S\ref{sec:ucr})? We show that models converge on accuracy while diverging on evidential grounding, with 69--97\% of correct predictions unsupported regardless of prompting strategy. Finally, we audit the measurement apparatus itself (\S\ref{sec:measurement}): agreement among judges is high but independence is low, perturbation tests fail, and reference choices shift 13\% of labels---yet the evidence gap is too large for any of these limitations to explain away, a scale argument that bounds random reference error but not systematic error.

\subsection{How Much Evidence Do Depression Dialogues Actually Contain?}\label{sec:landscape}

\begin{figure}[t]
\centering
\includegraphics[width=0.9\textwidth]{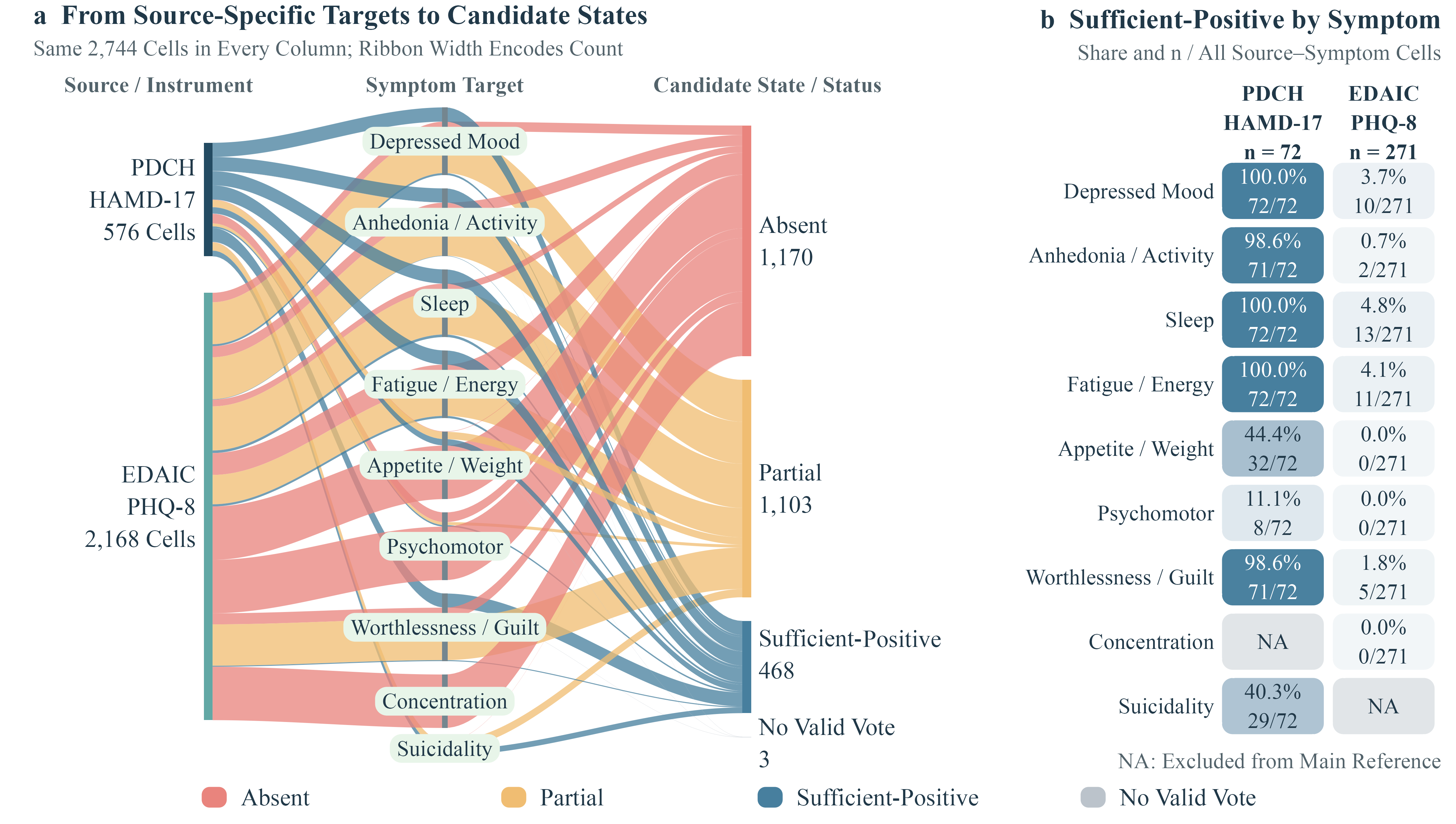}
\vspace{-15pt}
\caption{\textbf{Evidence landscape.}
\textbf{(a)}~Core collection; \textbf{(b,c)}~state proportions overall and by source; \textbf{(d)}~ribbon widths encode cell counts; \textbf{(e)}~sufficient-positive shares per source--symptom target (NA = excluded). All states are model-consensus candidates.}
\vspace{-15pt}
\label{fig:evidence-landscape}
\end{figure}

The fitted model-consensus reference assigns absent or partial evidence to \textbf{82.8\%} of cells (95\% CI: 82.1--83.5\%) across all 2,744 main cells 
: 42.6\% \absent{} ($n=1{,}170$), 40.2\% \partl{} ($n=1{,}103$), and only 17.1\% \suff{} ($n=468$), all with positive polarity. There are no sufficient-negative or conflicting MAP assignments.
The source-level contrast is stark: 74.1\% of PDCH cells receive sufficient evidence versus only 1.9\% of E-DAIC cells.
Source, language, instrument, and preprocessing are confounded, so this contrast is descriptive---but it underscores how heavily the evidence landscape depends on interview format (Figure~\ref{fig:within-source}). Figure~\ref{fig:evidence-landscape} shows the full distribution.
The candidate reference nonetheless retains the direction of the native labels while remaining far from them: sufficient evidence is assigned to 24.0\% of positive-native-label cells versus 8.3\% of negative-label cells, and mean native item scores increase monotonically across candidate states (absent 0.77, partial 0.91, sufficient 1.65); 100 sufficient-positive cells (3.6\%) carry negative native labels, so divergence is not one-sided.
Conditional on the native label, the lack of sufficient evidence is not confined to the negative side, where the negative-evidence contract is structurally harder (\S\ref{sec:taxonomy}): 76.0\% of positive-label cells lack sufficient evidence (95\% CI: 73.9--78.1\%), against 91.7\% of negative-label cells (95\% CI: 89.8--93.5\%; difference $-15.7$ points, 95\% CI: $-19.3$ to $-11.9$).
Because the contract makes positive sufficiency easier to reach, the positive-side share is, if anything, conservative; the label--support gap therefore cannot be attributed solely to the asymmetry of negative evidence.
Most assessment cells are evaluated against labels for which the dialogue provides no sufficient textual support. This scarcity is not a background condition---it directly determines the ceiling on evidence-grounded evaluation. What does a ``correct'' prediction on such cells actually reflect?

\subsection{Can Models Get the Right Label Without Evidence?}\label{sec:ucr}

\begin{figure}[t]
\centering
\includegraphics[width=0.96\textwidth]{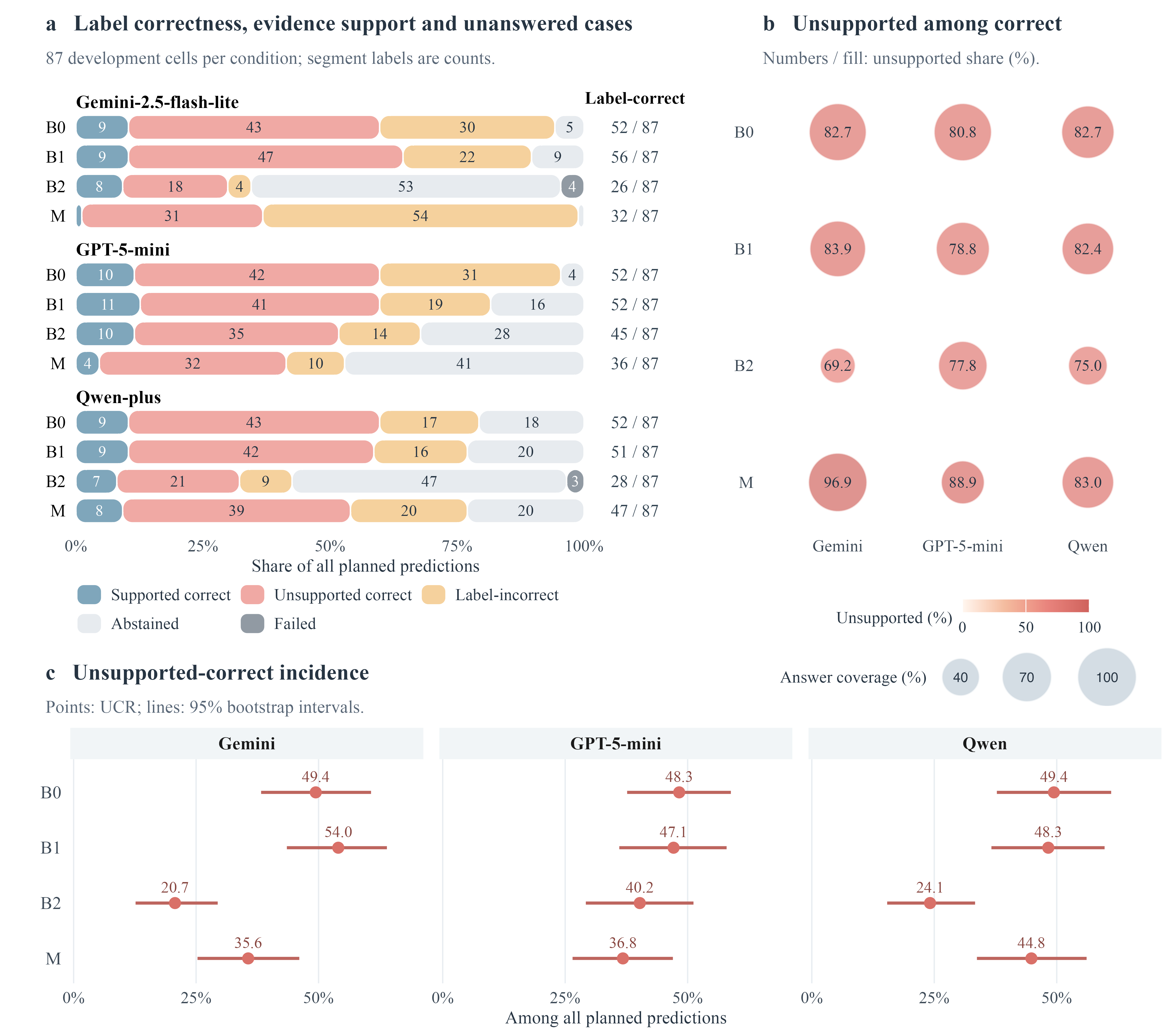}
\vspace{-15pt}
\caption{\textbf{Evidence-support audit across models and methods} (87 dev cells).
\textbf{(a)}~Five mutually exclusive outcomes; \textbf{(b)}~UCR with 95\% bootstrap intervals; \textbf{(c)}~hit caliber vs.\ coverage (circle area $=$ coverage).}
\vspace{-15pt}
\label{fig:model-evidence-audit}
\end{figure}

\noindent\textbf{The convergence paradox}
Under direct prompting (B0), all three models produce exactly $C = 52$ correct predictions out of 87 cells---yet their correct sets differ (intersection 42, union 64). Of the 42 jointly correct cells, only 8 have sufficient evidence aligned with the predicted label; 28 have partial or absent evidence, and 6 have sufficient evidence pointing the \emph{opposite} direction. Models converge on accuracy but diverge on evidential grounding.
The same $C = 52$ yields different apparent accuracy rates due to different answered denominators (Qwen 69 cells / 75.4\%, Gemini 82 / 63.4\%, GPT 83 / 62.7\%). N-based UCR is more consistent: .483--.494, with hit caliber at 80.8--82.7\% (Figure~\ref{fig:model-evidence-audit}). Approximately four out of five ``correct'' predictions lack evidence support regardless of which cells a model answers. If direct prompting already produces this pattern, can more sophisticated strategies resolve it?

\noindent\textbf{Can better prompting help?}
\textbf{B1} (evidence-first) requires a decisive quotation and binding verification before judging. UCR does not improve (.471--.540): requiring citation cannot overcome evidence scarcity. \textbf{B2} (sufficiency gate) achieves the lowest UCR (.207--.402) but at steep coverage cost (34.5--67.8\%), and 69.2--77.8\% of retained correct predictions still lack support. \textbf{M} (structured output) is the most model-dependent: Gemini abstains on 1.1\% of cells with 96.9\% hit caliber, while GPT abstains on 47.1\% with lower UCR (.368). Better prompting trades unsupported correctness for selective abstention without resolving the fundamental evidence scarcity (Table~\ref{tab:audit-and-measurement}).
A preregistered magnitude comparison cautions interpretation: hit-caliber estimates (0.692--0.969) bracket the reference-level not-sufficient prevalence (0.828), the pattern expected if model--method conditions differ mainly in their exposure to a shared label--support misalignment rather than in evidence use. Tier-level paired contrasts, which this grid reserves for a frontier-model comparison, remain unexecuted.

\begin{table*}[t]
\centering
\caption{
UCR and companion metrics on 87 development cells ($R=R_c=0$), 
together with the agreement and correlation structure of the legacy LLM judge panel.
Bracketed values are bootstrap 95\% CIs conditional on the provisional reference.
Failure is .000 except Gemini-B2 (.046) and Qwen-B2 (.034).
}
\label{tab:audit-and-measurement}

\scriptsize
\setlength{\tabcolsep}{2.4pt}
\renewcommand{\arraystretch}{0.94}

\begin{minipage}[t]{0.325\textwidth}
\centering
\textit{(a) Direct prompting (B0, B1)}\\[2pt]

\begin{tabular}{@{}llcccc@{}}
\toprule
Model & M. & UCR [CI] & Hit C. & Abst & Cov \\
\midrule
Gem. & B0 & .494 [.38,.61] & .827 & .057 & .943 \\
Gem. & B1 & .540 [.44,.64] & .839 & .103 & .897 \\
GPT  & B0 & .483 [.38,.59] & .808 & .046 & .954 \\
GPT  & B1 & .471 [.36,.58] & .788 & .184 & .816 \\
Qwen & B0 & .494 [.38,.61] & .827 & .207 & .793 \\
Qwen & B1 & .483 [.37,.60] & .824 & .230 & .770 \\
\bottomrule
\end{tabular}
\end{minipage}
\hfill
\begin{minipage}[t]{0.325\textwidth}
\centering
\textit{(b) Selective / structured (B2, M)}\\[2pt]

\begin{tabular}{@{}llcccc@{}}
\toprule
Model & M. & UCR [CI] & Hit C. & Abst & Cov \\
\midrule
Gem. & B2 & .207 [.13,.29] & .692 & .609 & .345 \\
Gem. & M  & .356 [.25,.46] & .969 & .011 & .989 \\
GPT  & B2 & .402 [.29,.51] & .778 & .322 & .678 \\
GPT  & M  & .368 [.27,.47] & .889 & .471 & .529 \\
Qwen & B2 & .241 [.15,.33] & .750 & .540 & .425 \\
Qwen & M  & .448 [.34,.56] & .830 & .230 & .770 \\
\bottomrule
\end{tabular}
\end{minipage}
\hfill
\begin{minipage}[t]{0.32\textwidth}
\centering
\textit{(c) Legacy judge panel quality}\\[2pt]

\setlength{\tabcolsep}{2pt}
\renewcommand{\arraystretch}{1.3}

\begin{tabular}{@{}lr@{\hspace{1em}}lr@{}}
\toprule
\multicolumn{2}{@{}c}{Agreement}
& \multicolumn{2}{c@{}}{Dependence} \\
\midrule
Fleiss $\kappa$ & .785
& PR (core) & 1.40 \\

Obs.\ agr. & .866
& PR (ext.) & 1.31 \\

Unanimous & 73.1\%
& Top eig.\ share & 84--87\% \\

&
& $E\rho^{2}(5)$ & .949 \\
\bottomrule
\end{tabular}
\end{minipage}
\vspace{-10pt}
\end{table*}

\subsection{How Robust Is the Evidence Measurement Itself?}\label{sec:measurement}

These findings rest on the model-consensus reference, whose robustness we now examine from three angles: inter-judge independence, perturbation sensitivity, and reference-method agreement.

\noindent\textbf{Agreement versus independence}
A legacy panel of five LLM protocols annotated 5,283 cells across 587 transcripts. Agreement is high: Fleiss' $\kappa = .785$, observed agreement $.866$, 73.1\% unanimous (Table~\ref{tab:audit-and-measurement}) (c). However, the eigenvalue participation ratio yields only \textbf{1.40}---five judges contribute roughly the information of 1.4 independent raters, with 84--87\% of eigenvalue mass on the first component. G-theory confirms: protocol variance is 1.0\% of total; $E\rho^2 = .949$ is high because judges barely disagree, not because each independently validates the evidence.


Despite low independence at the panel level, protocol choice matters for individual cells: P2 and P5 assign assessability at 3.3--3.4$\times$ the odds of P1 and P3 ($\chi^2 = 1{,}051.5$, $p < 10^{-200}$), and 47 of 57 tied cells change assignment under MFRM correction. An MTMM analysis yields convergent-validity ratio 2.39 (same-dimension $r = .738$ vs.\ different-dimension $r = .309$). On the new three-judge panel, an MTMM decomposition gives convergent 0.514 versus discriminant 0.235 (ratio 2.18; centered 0.360 vs.\ $-0.025$), with the language blocks closer than in the legacy panel (English 0.473, Chinese 0.425)---trait structure that is not explained by a judge-style factor, though convergent correlations are not independent validation. Full decompositions are in Appendix~\ref{app:gtheory}.

\noindent\textbf{Perturbation stress tests}
Eight perturbation classes spanning six failure modes test whether judges respond predictably to controlled evidence changes (gate: $\geq$85\% accuracy, $\geq$75\% posterior lower bound). No condition passes in either run (Table~\ref{tab:perturbation-results}). The highest hit rate is 20/61 (32.8\%). Initial failures included prompt--validator mismatches and provider errors; after repair, semantic accuracy remains below threshold.

\textbf{Where do missing sufficient-negative judgments come from?} Because the reference contains no sufficient-negative MAP assignments, we adjudicate the denial-insertion class (P3, expected \suff{}-negative) separately under the repaired instrument. The judge families split: Gemini returns sufficient-negative on 10/16 valid P3 judgments, Claude on 0/5; across all 269 valid anchor votes, 36 are sufficient-negative, against 9 under the initial run. The construct is thus producible by one judge family and absent from another under identical inputs, locating the main cause in judge capability rather than the construct being unreachable. A residual mismatch sits at the \confl{} boundary: on contradiction anchors (P7), Gemini resolves the conflict as denial overriding evidence (\suff{}-negative on 10/12 valid judgments) where the contract expects \confl{}, and Claude returns \confl{} on only 1/13. These counts are small, and Qwen's P3/P7 votes are missing, so the adjudication is directional.


\begin{table}[t]
\centering
\vspace{-10pt}
\caption{Perturbation anchors by failure mode. Entries are successes/total, no condition passes the gate.}
\label{tab:perturbation-results}

\scriptsize
\renewcommand{\arraystretch}{0.82}
\setlength{\tabcolsep}{2pt}
\setlength{\aboverulesep}{0.35ex}
\setlength{\belowrulesep}{0.35ex}

\begin{tabular*}{\columnwidth}
{@{\extracolsep{\fill}}cllccc@{}}
\toprule
Mode & Failure mode & Anchors & Claude & Gemini & Qwen \\
\midrule
F1 & Subject binding  & P4-other      & 2/11  & 0/11  & 0/11 \\
F2 & Temporal binding & P5, P4-ow, P8 & 12/64 & 16/64 & 0/64 \\
F3 & Assertion scope  & P4-wd, P3     & 1/35  & 10/35 & 0/35 \\
F4 & Withdrawal       & P2, P4-wn     & 4/36  & 6/36  & 0/36 \\
F5 & Aggregation      & P7, P1        & 6/50  & 2/50  & 2/50 \\
F6 & State projection & P6a, P6b, P8  & 17/61 & 20/61 & 0/61 \\
\bottomrule
\end{tabular*}
\vspace{-10pt}
\end{table}

\noindent\textbf{Reference sensitivity}\label{sec:sensitivity}
RE-LCA fusion covers 99.9\% of cells versus 77.3\% for plurality, but they disagree on 281/2,120 jointly labeled cells (13.3\%). Anchor validation does not favor fusion (67--72\% error vs.\ 55--64\% for plurality, McNemar $p < .02$). A one-hot mean baseline is identical to plurality unit by unit (2,744/2,744 cells): with at most three categorical votes, the mean judge degenerates to the modal judge, so reference construction chooses between abstention and model-based resolution of split votes, not between naive and sophisticated averaging. Plurality itself determines six sufficient-negative cells that fusion assigns none, so the zero sufficient-negative count is a property of the fusion fit rather than of every aggregator. A first entropy calibration attempt met the bin-occupancy criterion in three of five bins but fell at the support edge (223/246 voted views carry a single vote; error rates non-monotone in entropy), so no single-point unresolved threshold is adopted. Adaptation of an existing human-annotation subset yields 29 overlapping cells with agreement 27.6--79.3\% depending on an unresolved temporal-rule assumption. A separate blind 120-cell human-annotation study (\S\ref{sec:Human-validation}) confirms substantial agreement with both RE-LCA and plurality references and preserves the main evidence-gap findings under re-scoring.

\noindent\textbf{Robustness despite measurement limits}
The measurement apparatus is imperfect---but the evidence gap is too large to be an artifact. Moving cells from absent to sufficient at the disagreement rate (10--13\%) would still leave the majority unsupported. UCR values of .207--.540 across 12 conditions, with hit caliber consistently above 69\%, represent a pattern too pervasive for any reference choice to explain away. This scale argument bounds random reference error; it does not substitute for reference validity, because reference error could be systematic rather than random (\S\ref{sec:limitations}). The blinded human validation in \S\ref{sec:Human-validation} provides an independent check: re-scoring against the human reference preserves the main findings.

\vspace{-10pt}
\section{Human validation}\label{sec:Human-validation}
\vspace{-10pt}

\begin{figure}[t]
\centering
\includegraphics[width=1\textwidth]{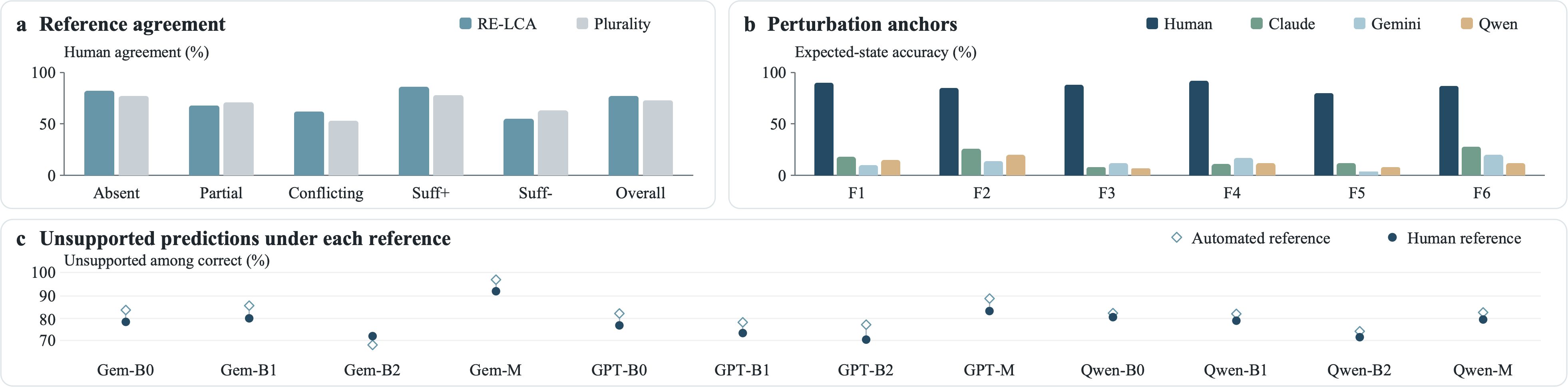}
\vspace{-15pt}
\caption{\textbf{Human validation.}
\textbf{(a)} RE-LCA and plurality show substantial agreement with blinded human consensus, with remaining disagreement concentrated near difficult evidence boundaries.
\textbf{(b)} Humans reliably recover the intended states of controlled perturbation anchors, whereas LLM judges do not.
\textbf{(c)} Re-scoring model predictions against the human reference preserves the high unsupported-among-correct rates across model--method conditions.}
\vspace{-15pt}
\label{fig:human-validation}
\end{figure}

Two blinded annotators independently labeled 120 assessment cells, with disagreements adjudicated by a third rater. RE-LCA showed 76.5\% overall exact agreement with human consensus, compared with 73.1\% for plurality, with lower agreement concentrated near the \textsc{Conflicting} and \textsc{Suff-} boundaries (Fig.~\ref{fig:human-validation}a). Humans also recovered the intended perturbation states at 80--92\% across six failure modes, whereas individual LLM judges achieved only 4--28\% (Fig.~\ref{fig:human-validation}b). Most importantly, re-scoring the overlapping model audit against the human reference preserved the main finding: unsupported-among-correct averaged 79.1\%, versus 81.8\% under the automated reference, with a mean absolute change of 3.1 percentage points and a maximum change of 4.9 points across 12 model--method conditions (Fig.~\ref{fig:human-validation}c).

\vspace{-10pt}
\section{Discussion \& Conclusion}\label{sec:discussion}
\vspace{-10pt}
Our results show that label correctness and textual support are distinct properties. Across \textsc{EviDial}, many assessment cells lack sufficient evidence in the observable dialogue, and many label-correct predictions remain unsupported across models, prompting strategies, and reference choices. This does not reveal how models reason internally; rather, it shows that accuracy alone cannot establish evidence-grounded assessment. The strong variation across sources further indicates that evidence support depends on the \emph{benchmark--instrument--transcript interface}, including how symptoms are elicited, mapped, and preserved in the evaluated text.
Benchmark evaluation should therefore treat evidence support as a separate axis alongside label accuracy. This distinction extends beyond depression dialogue whenever evaluation targets contain information not recoverable from the model-visible input. Although \textsc{EviDial} does not establish clinical validity and its references remain subject to measurement uncertainty, it makes a basic evaluation gap explicit: matching the benchmark label and being justified by the available evidence are distinct evaluation dimensions, and reliable behavioral assessment should measure both.

\vspace{-10pt}
\section{Limitations \& Future Directions}\label{sec:limitations}
\vspace{-10pt}
\textbf{Construct scope.}
\textsc{EviDial} measures whether the transcript provides sufficient textual evidence for a symptom assessment, not whether the patient clinically has that symptom. Its references therefore characterize evidence availability rather than diagnostic truth.
\textbf{Dataset composition.}
EDAIC contributes 79\% of the main reference cells, and language, source, and clinical instrument are confounded across datasets. Consequently, source-level differences cannot be uniquely attributed to any one of these factors.
\textbf{Model-audit scope.}
The development audit in \S\ref{sec:ucr} uses 87 cells, lightweight models, and single runs per condition. These experiments establish the presence of an evidence gap but are not intended to support definitive model rankings; broader held-out evaluation with stronger models and repeated runs remains future work.
\textbf{Reference validity.}
The automated reference remains imperfect: coverage is uneven, and some evaluator and judge families overlap. A blinded 120-cell human validation (\S\ref{sec:Human-validation}) confirms substantial agreement with the automated reference (76.5\% for RE-LCA) and shows that humans reliably recover perturbation anchor states (80--92\%) where LLM judges do not; however, this covers a subset of cells and does not constitute a complete audit across all dimensions. Negative evidence is also harder to establish under the current contract, making negative-direction estimates more sensitive to judge behavior. These limitations motivate treating the reference as a calibrated measurement instrument rather than definitive ground truth.

\clearpage
\section*{Ethics Statement}

This work uses secondary analysis of existing depression-dialogue corpora.
PDCH transcripts are licensed under CC BY-NC 4.0; EDAIC access is
request-gated through the original data holders; TalkDep transcripts are
publicly available. TalkDep profiles are simulated, whereas the main reference
uses clinical interview records. Public availability does not establish
uniform redistribution rights, and any release must respect the permissions
and access conditions of the original data sources.
The study additionally includes a blinded human-validation experiment
(\S\ref{sec:Human-validation}). Two independent annotators labeled 120
assessment cells under the four-state evidence taxonomy, with disagreements
adjudicated by a third rater. These annotations were used to evaluate the
agreement of the automated reference with human judgments and to test whether
the main audit conclusions were preserved under a human-derived reference.
The annotators did not provide clinical diagnoses; their task was restricted
to judging textual evidence under the predefined annotation contract.
A separate legacy annotation subset covers 99 PDCH interviews in a
suicidality-focused task. The available records for this legacy annotation do
not sufficiently document institutional approval, compensation, or support
arrangements to make claims about those procedures here.
The evidence-state annotations produced by EviDial are explicitly not
diagnostic labels: they characterize the availability and consistency of
textual evidence, not participants' underlying clinical status. In particular,
\absent{} or \partl{} should not be interpreted as evidence that a participant
does not have the corresponding symptom or condition; they indicate only that
the available transcript does not provide sufficient evidence under the
specified evidence contract.
\section*{Reproducibility Statement}

The accompanying archive contains model outputs, usage records, prompts, and
analysis scripts. Audited model responses predate the current reference export
and were retrospectively rescored. Fixed seeds identify the bootstrap and
fusion runs, and input hashes support artifact matching.
For the human validation (\S\ref{sec:Human-validation}), the paper specifies
the sampled assessment cells, blinded independent annotation procedure,
third-rater adjudication, agreement computation, and rescoring of the
overlapping model audit against the human-derived reference. The associated
analysis code reproduces the reported agreement statistics, perturbation-state
recovery results, and condition-level changes under human rescoring.
The endpoint runner checks only that two supplied freeze strings are nonempty;
it does not enforce a complete hash chain or chronological freeze order.
Legacy numerical errors and incomplete recovery checks are documented in
Appendix~\ref{app:lca}; invalid posterior summaries and absolute information
criteria are excluded as evidence. Cross-platform bit-identical reproduction
has not been established. Release of raw transcripts and annotation materials
derived from restricted corpora remains subject to the permissions of the
original data sources.
Code, analysis scripts, and reproducibility materials are available in an anonymized repository at \url{https://anonymous.4open.science/r/evidial-code-F024/README.md}.

\section*{AI Use Statement}

Generative AI tools were used to assist with language editing and polishing,
and to support literature retrieval and discovery of potentially relevant
prior work. All AI-assisted text was reviewed and revised by the authors,
and all references and factual claims obtained with AI assistance were
independently verified against the original sources. LLM judges evaluated
as part of the experiments were study subjects rather than sources of the
human-validation reference. The authors take full responsibility for the
final content of this work.

\bibliography{references}
\bibliographystyle{iclr2027_conference}

\clearpage
\appendix
\section{Source Corpus Details}\label{app:sources}

Table~\ref{tab:corpus-detail} provides descriptive statistics for the three source corpora, recomputed from the processed unified input.

\begin{table}[h]
\centering
\caption{Corpus descriptive statistics. A segment is one stored speaker-turn record, not a full clinician--patient exchange; score ranges are observed rather than theoretical.}
\label{tab:corpus-detail}
\small
\begin{tabular}{@{}lrrrrr@{}}
\toprule
Source & Transcripts & Median segments & Segment range & Median total & Total range \\
\midrule
PDCH & 99 & 276 & 159--642 & 15 & 1--35 \\
EDAIC & 271 & 60 & 21--129 & 6 & 0--23 \\
TalkDep & 12 & 65.5 & 34--73 & 18.5 & 5--40 \\
\bottomrule
\end{tabular}
\end{table}

PDCH contains real psychiatric diagnostic interviews conducted at a Chinese hospital, with 72 unique participants (some interviewed more than once). Its 37 reconstructed totals are sums of item scores. EDAIC consists of semi-structured distress interviews from the Extended Distress Analysis Interview Corpus, scored on the PHQ-8. TalkDep uses LLM-simulated patient personas following BDI-II profiles. The 370 clinical transcripts represent 343 participants; the other 12 transcripts represent 12 virtual personas.

For comparison, the legacy diagnostic pool contains 587 transcripts: PDCH 99, EDAIC 271, TalkDep 12, D4 60, LingxiDiag 52, MDD5K 51, and EATD 42. All nine dimensions enter that analysis, yielding $587\times9=5{,}283$ cells. These are distinct denominators from the current main reference and model-output sample.

\section{Generalizability Theory and MFRM Details}\label{app:gtheory}

\paragraph{G-theory model.}
We decompose vote-level observations $y_{upk}$ for participant $p$, unit (cell) $u$ nested in $p$, and protocol $k$:
\begin{equation}
y_{upk} = \mu + \pi_p + \nu_{u:p} + \tau_k + \varepsilon_{upk}
\end{equation}
where $\pi_p$ and $\nu_{u:p}$ are mean-zero random effects, $\tau_k$ are fixed protocol effects, and $\varepsilon_{upk}$ is residual error confounding the unit$\times$protocol interaction with pure error (single observation per cell-protocol combination). Here $y_{upk}$ measures binary assessability, and the participant grouping includes source-recorded virtual or synthetic identities. The legacy sample comprises 5,283 cells from 587 transcripts and seven sources, with 560 source-recorded identities. The 99 PDCH interviews correspond to 72 participants, with 27 participants contributing two interviews.

Variance components are estimated via method-of-moments (Type-I SS) with nested expected-mean-square coefficients. Protocol mean effects are removed before the nested decomposition, and the protocol component is the observed variance across the five protocol means. For measure A (all votes, $n = 26{,}415$):
$\hat{\sigma}^2_\pi = 0.054$, $\hat{\sigma}^2_{\nu:p} = 0.131$, $\hat{\sigma}^2_\tau = 0.002$, $\hat{\sigma}^2_\varepsilon = 0.050$.

The relative generalizability coefficient for $k = 5$ fixed protocols:
\begin{equation}
E\rho^2(5) = \frac{\hat{\sigma}^2_\pi + \hat{\sigma}^2_{\nu:p}}{\hat{\sigma}^2_\pi + \hat{\sigma}^2_{\nu:p} + \hat{\sigma}^2_\varepsilon / 5} = 0.949
\end{equation}
This calculation treats residual contributions as exchangeable under replication of the fixed protocol set. The secondary absolute-decision coefficient also includes the observed protocol component in the error term and equals $.946693$. Neither coefficient supplies an external criterion for the correctness of a cell's evidence label.

\paragraph{MFRM calibration.}
The Many-Facet Rasch Model estimates protocol severity parameters via conditional maximum likelihood. For each binary submeasurement, the Rasch specification is $\operatorname{logit}P(y_{uk}=1\mid\theta_u,\beta_k)=\theta_u-\beta_k$, with $\sum_k\beta_k=0$. The assessability fit uses all 26,415 votes; the conditional-direction fit uses 16,216 observable votes, with 1,508 all-not-assessable cells contributing no direction observations. Protocol log-odds estimates describe assessability: P2 and P5 have odds multipliers of 3.3--3.4$\times$ relative to a zero-severity protocol; P1 and P3 have multipliers of 0.34--0.38$\times$ for an assessable category. The conditional likelihood ratio test for protocol homogeneity yields $\chi^2 = 1{,}051.5$ ($df = 4$, $p < 10^{-200}$). The assessability odds multipliers $\exp(-\beta_k)$ for P1--P5 are $.3431$, $3.2673$, $.3766$, $.6905$, and $3.4313$, respectively. The corresponding conditional-direction multipliers are $1.1539$, $.9959$, $1.3493$, $.4415$, and $1.4606$.

Frozen plurality yields 2,378 positive, 869 negative, and 2,036 not-assessable assignments. MFRM yields 2,409, 875, and 1,999, respectively. Of 57 tied cells under plurality, 47 change their assignment under MFRM calibration. These differences quantify disagreement between aggregation procedures. Against a selected 45-cell model re-adjudication, both plurality and MFRM agree on 27 cells; against two model re-readers on an 18-cell set, both agree on 13 cells per reader. Among 22 uniquely joinable cells from a separate annotation pilot, agreement with the mapped pilot plurality is $13/22$ for legacy plurality and $12/22$ for MFRM. These small comparisons involve selection and cross-version mappings and do not establish reference-quality improvement.

\section{Legacy RE-LCA Analyses and Recovery Checks}\label{app:lca}

The random-effects latent class model assigns each cell $u$ a latent legacy assessment category $Z_u\in\{\text{observable-positive},\text{observable-negative},\text{not-assessable}\}$ with a cell-level random effect on protocol emissions. The exploratory models are fitted directly to the legacy three-category votes; the covariate extension changes class priors through additive source and dimension-group terms. MFRM assignments are loaded afterward for comparison. This analysis is separate from construction of the current four-state reference.

The archived implementation has two numerical limitations relevant to interpretation. Its posterior summary function rescales each class separately over integration points without preserving the class-specific scaling factors; the resulting posterior probabilities, entropy, MAP summaries, and ICL values are therefore excluded from our evidence. Its discrete integration also omits a constant grid-width factor when reporting likelihoods, affecting absolute likelihood and information-criterion values. We do not use the archived absolute AIC/BIC values as model-selection evidence. The current reference implementation is evaluated separately; the legacy results cannot validate it by themselves.

\paragraph{Synthetic parameter recovery.}
The available recovery experiments comprise five scenarios, each using one simulated dataset of 5,283 cells. They assess parameter error against declared tolerances. Three scenarios concern the homogeneous model: realistic signal and zero random-effect variance meet their parameter tolerances; a low-signal scenario fails, with maximum class-prevalence error $.1029$ against tolerance $.05$, maximum population-emission error $.1241$ against $.07$, and random-effect standard-deviation error $.3228$ against $.30$.

Two additional scenarios concern the covariate extension. A crossed source--language design meets its recovery criteria. A scenario preserving the observed confounded source structure fails the source-specific prevalence criterion: maximum error is $.0603$ against tolerance $.03$. These checks provide limited evidence about parameter recovery in selected simulated settings. They do not estimate repeated-simulation class-assignment accuracy.

The legacy fitted-model simulation checks also show incomplete fit: the homogeneous specification fails the recorded stratified criterion in 15 of 16 strata; the source-plus-dimension-group extension fails in 11 of 16; and an exploratory source-plus-nine-dimensions extension fails in three of 16. These results limit substantive interpretation of the legacy latent-class models and their use for reference validation.

\section{MTMM and Language Comparisons}\label{app:mtmm}

\paragraph{Multitrait--Multimethod analysis.}
Following \citet{campbell1959mtmm}, we compute mean same-trait, different-method correlations and mean different-trait, same-method correlations for legacy binary assessability. Results: same-trait correlation $r = 0.738$, different-trait correlation $r = 0.309$, ratio $= 2.39$. After participant-mean centering: the corresponding correlations are $r = 0.620$ and $r = -0.069$. Centering changes the correlation structure, and the negative value should be interpreted accordingly. These descriptive correlations do not independently confirm construct validity.

The matrix uses nine symptom dimensions as traits and five protocols as methods. Repeated observations for the same identity, dimension, and protocol are averaged, giving 560 rows and 45 variables. The mean same-trait, different-method correlation averages 90 pairs; the different-trait, same-method mean averages 180 pairs. Within the Chinese-language block, the uncentered correlations are $.8404$ and $.5118$ across 277 identities; within the English-language block, they are $.5479$ and $.0538$ across 283 identities. The Chinese block combines PDCH, D4, EATD, LingxiDiag, and MDD5K; the English block combines EDAIC and TalkDep.

\paragraph{Spectral summary.}
For correlation-matrix eigenvalues $\lambda_1,\ldots,\lambda_5$, the participation ratio is $(\sum_k\lambda_k)^2/\sum_k\lambda_k^2$. Its interpretation concerns the raw vote vectors. For example, five raters that independently flip a balanced binary true label with probability $.05$ each have marginal pairwise vote correlation $.81$. Their population participation ratio is $5/(1+4\times .81^2)=1.380$. Thus a low participation ratio can arise from shared signal even with conditionally independent errors. Assessing error dependence requires additional assumptions or an external reference.

\paragraph{Language comparison.}
The legacy language analysis was a descriptive parameter screen using separately fitted binary assessability models and 25 parametric bootstrap datasets per block. Source, instrument, language, and sample provenance are confounded. Moreover, one flagged contrast involved random-effect standard deviations whose reported bootstrap standard errors were zero under a discrete parameter search. We therefore make no formal measurement-invariance claim from that screen and do not interpret its estimated assessability prevalence as symptom-positive prevalence or as a causal language effect.

\section{Perturbation Class Definitions}\label{app:perturbations}

The eight perturbation classes and their expected final states are given in Table~\ref{tab:perturbation-classes}. These are the construction targets used by the executed scoring scripts, not independently verified labels for the original interviews.

\begin{table}[h]
\centering
\caption{Perturbation classes with expected evidence-state targets. P4 inserts distractors while preserving the original supporting evidence; its four variants test invariance under each binding operation.}
\label{tab:perturbation-classes}
\small
\begin{tabular}{@{}llll@{}}
\toprule
Class & Manipulation & Expected final state & Failure mode \\
\midrule
P1 & Deletion (evidence removed) & \absent{} & F5: Aggregation \\
P2 & Withdrawal (later retraction) & \absent{} & F4: Withdrawal \\
P3 & Denial insertion & \suff{}-neg & F3: Assertion scope \\
P4-ow & Out-of-window distractor & \suff{}-pos & F2: Temporal \\
P4-op & Other-person attribution & \suff{}-pos & F1: Subject \\
P4-wd & Weak denial & \suff{}-pos & F3: Assertion scope \\
P4-wn & Withdrawn statement & \suff{}-pos & F4: Withdrawal \\
P5 & Temporal boundary test & \absent{} & F2: Temporal \\
P6a & Half-support reduction & \partl{} & F6: State projection \\
P6b & Partial evidence only & \partl{} & F6: State projection \\
P7 & Contradiction insertion & \confl{} & F5: Aggregation \\
P8 & Window-incomplete denial & \partl{} & F2/F6: Temporal/Projection \\
\bottomrule
\end{tabular}
\end{table}

The initial run contains 319 views; the update contains 343. The initial scoring script treats P6b as descriptive, while the updated script includes it in the gated set. The reported F1--F6 groups aggregate these variants and are not independent because P8 appears twice. Scoring tests agreement with the expected final state rather than an estimated original-to-perturbed causal effect. Human assessment of perturbation anchors was completed as part of the blinded validation study (\S\ref{sec:Human-validation}), with humans recovering intended states at 80--92\% across failure modes. Generator-family exclusivity ensures the model generating perturbations is from a different family than the judges evaluating them.

\section{Reference Fusion Specification}\label{app:reference-fusion}

Let $Y_{uj}$ be the categorical vote of judge $j$ on cell $u$, and let $\mathcal O_u$ contain only judges with valid votes on that cell. Write $\pi_k(\boldsymbol x_u)$ for the five-state softmax prevalence model and $e_j(y\mid k,b)$ for the categorical emission probability. The random effect $b_u$ is shared by judgments on the same cell; effects for different cells are integrated separately. Using a zero-mean Gaussian density $\phi_\sigma$ restricted to the numerical integration interval, the observed-vote likelihood is
\begin{equation}
\mathcal L(\theta)
=\prod_{u:\,|\mathcal O_u|>0}
\sum_{k=1}^{5}\pi_k(\boldsymbol x_u)
\int_{-6}^{6}
\left[\prod_{j\in\mathcal O_u} e_j(Y_{uj}\mid k,b)\right]
\frac{\phi_\sigma(b)}{\int_{-6}^{6}\phi_\sigma(t)\,dt}\,db.
\label{eq:reference-fusion-likelihood}
\end{equation}
The design vector contains an intercept, a joint source/language indicator, and a dimension-group indicator, without an interaction term. Computation replaces each integral with a sum over $b_g=-6+0.15g$, $g=0,\ldots,80$, using normalized Gaussian weights
\begin{equation}
w_g(\sigma)=
\frac{\exp[-b_g^2/(2\sigma^2)]}
{\sum_{h=0}^{80}\exp[-b_h^2/(2\sigma^2)]}.
\end{equation}
Thus, $b_u$ is represented by a finite-grid approximation to $\mathcal N(0,\sigma^2)$, not a participant-level intercept. The fit additionally incorporates the emission pseudo-counts described in \S\ref{sec:reference}. Posterior state probabilities sum jointly over the grid and are normalized across states.

The production likelihood sum omits a constant grid-width factor, so its stored log likelihood differs from the normalized-grid expression by $-n\log(0.15)$ for $n$ voted cells. This constant does not change posterior assignments or comparisons on the same grid and sample. Absolute stored likelihood or information-criterion values are not reported as evidence here. The current posterior routine preserves the relative scaling across states, unlike the legacy routine discussed in Appendix~\ref{app:lca}.

\end{document}